\def\year{2022}\relax
\documentclass[letterpaper]{article} 
\usepackage{aaai22}  
\usepackage{amsmath}
\usepackage{amssymb}
\usepackage{times}
\usepackage{helvet}
\usepackage{courier}
\usepackage{booktabs}
\usepackage{textcomp}
\usepackage{xcolor}
\usepackage{times}  
\usepackage{helvet}  
\usepackage{courier}  
\usepackage[hyphens]{url}  
\usepackage{graphicx} 
\usepackage{multirow}
\usepackage{subcaption}
\usepackage{tabularx}

\usepackage{natbib}  
\usepackage{caption} 
\DeclareCaptionStyle{ruled}{labelfont=normalfont,labelsep=colon,strut=off} 
\usepackage{algorithm2e}
\usepackage{float}

\usepackage{newfloat}
\usepackage{listings}
\title{Domain Agnostic Few-Shot Learning For Document Intelligence}
\author{\vspace{-70cm}}
\author {
 Jaya Krishna Mandivarapu,
     Eric Bunch,
    Glenn Fung
}
\affiliations {
        American Family Insurance Research\\

  jmandivarapu1@student.gsu.edu, ebunch@amfam.com, gfung@amfam.com
}

\usepackage{bibentry}

\begin{document}
\maketitle
\begin{abstract}
 Few-shot learning aims to generalize to novel classes with only a few samples with class labels. Research in few-shot learning has borrowed techniques from transfer learning, metric learning, meta-learning, and Bayesian methods. These methods also aim to train models from limited training samples, and while encouraging performance has been achieved, they often fail to generalize to novel domains. Many of the existing meta-learning methods rely on training data for which the base classes are sampled from the same domain as the novel classes used for meta-testing. However, in many applications in the industry, such as document classification, collecting large samples of data for meta-learning is infeasible or impossible. While research in the field of the cross-domain few-shot learning exists, it is mostly limited to computer vision. To our knowledge, no work yet exists that examines the use of few-shot learning for classification of semi-structured documents (scans of paper documents) generated as part of a business workflow (forms, letters, bills, etc.).  Here the domain shift is significant, going from natural images to the semi-structured documents of interest. In this work, we address the problem of few-shot document image classification under domain shift. We evaluate our work by extensive comparisons with existing  methods. Experimental results demonstrate that the proposed method shows consistent improvements on the few-shot classification performance under domain shift.

\end{abstract}
\section{Introduction}
The challenges of document classification in an industry setting are many: scalability, accuracy and degree of automation, speed of delivery requirements, and limited time available by domain business experts. While a company may have a large collection of documents and high level metadata consisting of broad classes spanning this collection, there are inevitably use cases and workflows that need more granular and specialized document classes. The document classification within these specialized workflows is typically done by business and domain experts, whose time is valuable, and better spent on other tasks. This setup calls for an automation process that can be trained to classify document sub-classes with smaller amounts of training data. Few-shot learning methods offer exactly this benefit. 

One possible approach is to pre-train a model on available large open-source document datasets. However, they tend to be significantly different from the internally generated workflow documents. Some reasons for this are that the open source document data sources may be scanned using lower quality scanners, the documents are of different types than those considered internal to the company, or the text content itself utilizes a different colloquial vocabulary. This paper proposes a few-shot meta-learning technique that utilizes both the visual and text components of a document, and is pre-trained on open source document datasets that are out of domain with respect to the internal company documents, which the model is evaluated on.

\section{Related Work}
This work explores a method which can be used for few-shot learning on multi-channel document data, in which the meta-training is done on a distinct domain of open source documents. Work has been done in this area addressing the separate problems of: 
    \begin{enumerate}
        \item Meta-training a few-shot model on document data,
        \item Combining visual and textual feature channels via canonical correlation, and
        \item Domain adaptation of models trained on image data.
    \end{enumerate}
This work proposes to address all of these issues with a single approach, driven by a distinct business need. Here we detail previous work done in each of these directions.

\subsection{Meta-learning} Meta-learning has been a powerful tool to answer the challenge of the large data requirements that many deep learning models seem to face. So far, many of the applications of deep meta-learning have been in few-shot image classification \cite{finn2019online,snell2017prototypical,ravi2016optimization}. Meta-learning for few shot image classification has often been evaluated on data sets such as ImageNet \cite{russakovsky2015imagenet}, CIFAR-10 and CIFAR-100 \cite{Krizhevsky2009LearningML}, and Omniglot \cite{Lake2011OneSL}. However, there has been little done to apply the methods of meta-learning to industry level document images. 

\subsection{Canonical Correlation} One aspect of enterprise level document images is that they typically have two feature channels; a visual channel, and a text channel. Each has useful information that can be leveraged for document classification. However, a challenge to overcome with this is that typically pre-trained models are used as feature extractors, which are further fine tuned during meta-training. The vectors extracted by these pre-trained models (one for each channel mentioned above) are typically not the same length, and encode the information of the channel in semantically different ways. One method of overcoming this challenge is the use of Canonical Correlation \cite{hotelling1992relations,akaho2006kernel, melzer2001nonlinear,bach2002kernel,andrew2013deep}, which in a sense aligns two vectors via projections in such a way that the projections are maximally correlated. We use a later iteration of this method called Deep Canonical Correlation \cite{andrew2013deep}. This allows us to efficiently combine the two modalities of visual and text features occurring in enterprise document images for the purpose of few-shot meta-learning. 
\subsection{Domain adaptation}
In the traditional machine learning setting, the data samples used for training and testing an algorithm are assumed to come from the same distribution. In practical applications however, this is not always a valid assumption; the data available for training may fall into a different distribution than the data the model is expected to perform on in a live system. A typical example of this is a model which is trained on an open source data set is then desired to be used for inference in a smaller, proprietary data set, perhaps for a slightly different task. Domain adaptation is a subfield of machine learning that attempts to overcome this challenge. Typical approaches include transfer learning \cite{Pan2010ASO}, semi-supervised learning \cite{semisupervisedSurvey,semisupervisedSurvey_2}, multi-task learning \cite{Caruana2004MultitaskL}, and meta-learning \cite{Huisman_2021}. Recent methods include 
\citep{liu2020feature} which uses feature transformation ensemble model with batch spectral regularization, \citep{cai2020cross} aims train specific layers of network using MAML,
 \citep{jiang2020transductive} uses a new prediction head and global classification network based on semantic information for addressing the cross-domain adaptation.

\section{Methodology}

In this section, we briefly formally define the few-short learning and cross-domain few shot learning problems. We then describe our proposed method, Cross Domain Few-Shot Learning using  \textbf{D}eep \textbf{C}anonical  \textbf{C}orrelation for  \textbf{D}ocument  \textbf{I}ntelligence dubbed as  \textbf{DCCDI} in later parts of the paper.  \\
\begin{figure*}[t]
  \begin{center}
    \includegraphics[width=0.9\textwidth]{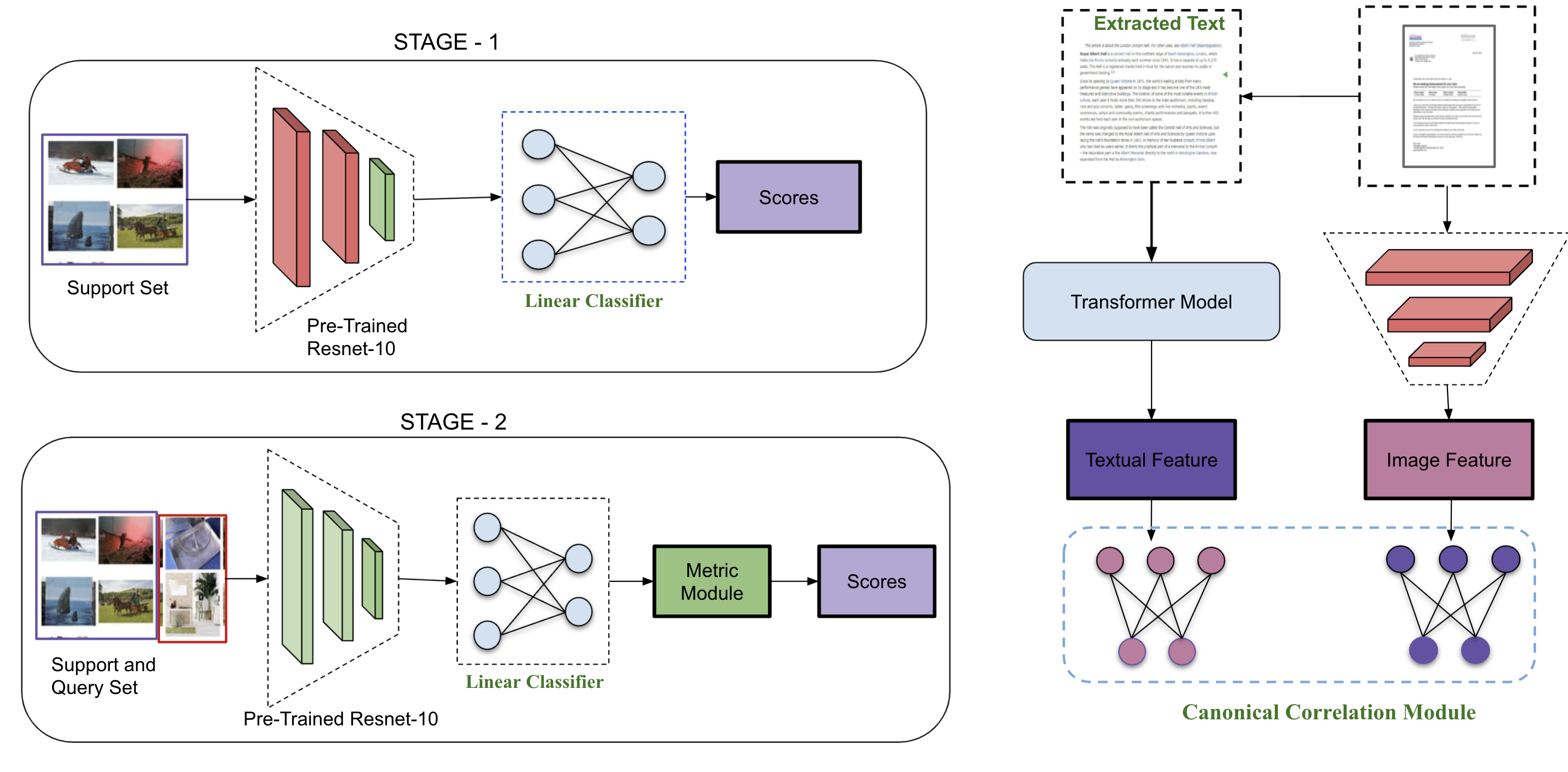}
  \end{center}
  \caption{The overall architecture of our approach. LEFT(Stage-1): Using episodic training paradigm train the last $k$ layers of ResNet-10 (shown in green color) using Cross-entropy loss by support set . Left(Stage-2): Using episodic training paradigm train all layers of ResNet-10 using Cross-entropy loss by support, Query set and by including the Metric-Learning Module. During Meta-testing all the Resnet-10 layers will be fixed. RIGHT: Canonical Correlation Block: During Meta-testing all the text extracted from document image and both image, textual features was trained using canonical correlation loss }
  \label{fig:overview}
\end{figure*}
\subsection{Formal problem definition}
Formally, a few-shot learning problem is denoted as $P=$ $( \mathcal{D}_{\text {source}}, \mathcal{D}_{\text {target}})$; $\mathcal{D}_{\text {source}}$ is  the meta-train set from where base classes as sampled for episodic training. Novel classes during meta-test are sampled from the target domain $\mathcal{D}_{\text {target}}$, such that $D_{\text {source}} \cap D_{\text {target }}=\emptyset$. For brevity, we will define the \textit{\textbf{domain}} of the source dataset $\mathcal{D}_{source}$ to be 

\begin{equation}
  \label{eqn:dist}
 d_{source}=\{\mathcal{X}, \mathcal{Y}, P_{source}\}
\end{equation}
\\
where $\mathcal{X}$ is the feature space of all the inputs in d-dimensional space; $\mathcal{X} \subset \mathbb{R}^{d} $ and $\mathcal{Y}$ is the label space of all the labels;  $\mathcal{Y} \subset  \{1, \ldots C\}$ where is the number of classes,  and  $P_{source}$ is the joint probability distribution over the feature,label pairs of  $\{ \mathcal{X}, \mathcal{Y}\}$ denoted by $p(x,y)$. A similar definition can be made for the target dataset.

We focus on few-shot settings for document classification using a model $f$ with parameters $\theta$, wich we will denote $f_{\theta}$, via meta learning tasks using episodic training from the meta-train set $\mathcal{D}_{source}$ and aim to demonstrate generalization to novel classes present in the meta-test set  $\mathcal{D}_{target}$. 

During meta training the model $f_{\theta}$ is provided with a wide range of classification tasks  $\mathcal{T}_i$ drawn from the dataset $\mathcal{D}_{\text {source}}=\{\mathcal{T}_{1}, \ldots, \mathcal{T}_{n}\}$ where each episodic task is $\mathcal{T}_{i}=\{(x_{1}^{i}, y_{1}^{i}), \ldots,(x_{k}^{i}, y_{k}^{i})\}$, and where ${x}_{i}$ represents image $i$ and ${y}_{i}$ its corresponding label. 

Each task $\mathcal{T}_i$ is further partitioned into a \textit{support set} $\mathcal{S}_i$ used for training, and a  \textit{query set} $\mathcal{Q}_i$ used for testing. That is, $\mathcal{T}_i$ can be written as the disjoint union $\mathcal{T}_i = \mathcal{S}_i \dot{\cup} \mathcal{Q}_i$. Overall, $\mathcal{D}_{source}$ can be written as  $\mathcal{D}_{\text {source}} = \{ (\mathcal{S}_1, \mathcal{Q}_1), \ldots, (\mathcal{S}_n, \mathcal{Q}_n)  \}$.

We follow the conventional way of preparing the support and query sets for each task ($\mathcal{T}_i$), which is a $C$-way, $N$-shot classification problem in which $C$ classes are randomly drawn from the entire set of classes from $\mathcal{D}$. Furthermore, for each of the sampled classes, $N$ and $M$ examples are sampled for the support and query set respectively such that each task $\mathcal{T}_i$ consists of $(\mathcal{S}_i, \mathcal{Q}_i)$ where $\mathcal{S}_i ={\{(\mathbf{x}_{i}, y_{i})\}}_{i=1}^{C \times N}$ is a support set consisting of $N$ labeled images for each of the $C$ classes and the query set $\mathcal{Q}_i = {\{\tilde{\mathbf{x}}_{i}, \tilde{y}_{i}\}}_{i=1}^{C \times M}$ with $M$ samples per class and $y, \tilde{y} \in\{1, \ldots, C\}$ are the corresponding class labels. 

The cross-domain few-shot learning scheme matches closely with our real-world industry setting where the source domain $\mathcal{D}_{source}$ and the target domain $\mathcal{D}_{target}$ belong to different distributions. As in   Eq. \ref{eqn:dist}, the joint distribution of the source dataset is indicated as $P_{source}$ and the target domain distribution can be denoted as $P_{target}$. Furthermore, as is the case in a  cross-domain few-shot learning setting,  $P_{source} \neq P_{target}$ and $\mathcal{Y}_{s}$ is disjoint from $\mathcal{Y}_{t}$. Also, similarly to a few-shot learning paradigm, during the episodic meta-training phase, the model is trained on a large number of tasks $\mathcal{T}_i$ sampled from the source domain $\mathcal{D}_{source}$. 

During the meta-testing phase, the model is presented with a support set $S=\left\{x_{i}, y_{i}\right\}_{i=1}^{K \times N}$ consisting of $N$ examples from $K$ novel classes and  $\mathcal{Q}=\{x_{i}, y_{i}\}_{i=1}^{K \times M}$ consisting of $M$ examples  which are very different from the meta-training classes.

After the meta-trained model $\mathnormal{\hat{f}_\theta}$ is adapted to the support set, a query set from novel classes is used to evaluate the model performance.


\subsection{Canonical Correlation}\label{tab:canonicalCorelation}
Ideally, an effective document classification method needs to leverage both textual and image (including layout) information. When using deep convolutional neural networks for document image classification, the document is treated as an image and is ingested as tensor representing the pixel values of the image to get the visual feature vector of the document images.On the other hand, all the text from the document image is extracted, converted into tokens and passed through a BERT-like pre-trained transformer-based language model to obtain textual features. Some of the ways of utilizing both the textual and visual features during the classification are to concatenate or average them before passing them through the final classification layer. Some of the disadvantages of doing this are a) More computational resources are required for model training for large dimensional features b) Difficult to maintain the synchronization between both the visual and textual modalities, which might impact model performance.

We address this dilemma by introducing the Deep Canonical Correlation for Document Intelligence Module (DCCDI) during meta-test phase to represent a document utilizing both the textual and visual features. By using the proposed DCCDI module, we produce highly correlated transformations of multiple modalities of data such as textual and visual using complex non-linear transformations. Canonical correlation was proposed by Hotelling   \citep{hotelling1992relations}. It is a widely used technique in the statistics community to measure the linear relationship between two multidimensional variables, used in finding linear projections of two multidimensional vectors that are maximally correlated. Later on it was applied to machine learning by different researchers \citep{akaho2006kernel, melzer2001nonlinear,bach2002kernel,andrew2013deep}. We use deep canonical corelation method propsed by \citep{andrew2013deep} in our DCCM module with the goal of achieving fine-grained cross-modality alignment between the visual and textual modalities. 

As shown in RIGHT Figure \ref{fig:overview}; $V \in \mathbb{R}^{N \times d_{1}}$ is the multidimensional vector for the visual modality where $d_{1}$ is total number of dimensions and $T \in \mathbb{R}^{N \times d_{2}}$ is the multidimensional vector for the textual modality where $d_{2}$ is total number of dimensions. $N$ is the total number of inputs available in each modality. The input multidimensional vectors in different modalities are transformed using two neural networks $\textsl{g}$ with parameters $\phi_{1}$, $\textsl{h}$ with parameters $\phi_{2}$

\begin{equation}
  \label{eqn:cca}
 \mathcal{Z}_{1}=\textsl{g}_{\phi_1} (V),
 \hspace{6pt}
  \mathcal{Z}_{2}=\textsl{h}_{\phi_2} (T)
\end{equation}

$\mathcal{Z}_{1} \in \mathbb{R}^{N \times d}$ and $\mathcal{Z}_{2} \in \mathbb{R}^{N \times d}$ are the $\mathnormal{d}$ dimensional outputs of the neural networks. Both the neural networks $\textsl{g}$, $\textsl{h}$ are optimized jointly with a goal of making the correlation between $\mathcal{Z}_{1}$ and $\mathcal{Z}_{2}$ as high as possible:

$$
(\phi_{1}^{*}, \phi_{2}^{*})
=
\underset{\phi_{1}, \phi_{2}}{\arg \max } \operatorname{corr}(\textsl{g}_{\phi_1}(V), \textsl{h}_{\phi_2}(T))
$$

The above equation can be solved in multiple ways. For this work we chose an approach suggested by \citep{martin1979multivariate}  and that utilizes Singular Value Decomposition.
Define the centered output matrices by $\bar{\mathcal{Z}}_{i}=\mathcal{Z}_{i}^{\prime}-\frac{1}{N} \mathcal{Z}_{i}^{\prime} \mathbf{1}$. Then define

\begin{align}
    \hat{\Sigma}_{11} 
    &= 
    \frac{1}{d-1} \bar{\mathcal{Z}}_{1} \bar{\mathcal{Z}}_{1}^{\prime}+r_{1} \\ \nonumber
    \hat{\Sigma}_{22} 
    &= 
    \frac{1}{d-1} \bar{\mathcal{Z}}_{2} \bar{\mathcal{Z}}_{2}^{\prime}+r_{1} \\ \nonumber
    \hat{\Sigma}_{12}
    &=
    \frac{1}{d-1} \bar{\mathcal{Z}}_{1} \bar{\mathcal{Z}}_{2}^{\prime} \\ \nonumber
\end{align}

\noindent where $r_1 > 0$ is a regularization constant. As discussed in \cite{andrew2013deep}, the correlation of the top $k$ components of $\mathcal{Z}_{1}$ and $\mathcal{Z}_{2}$ is the sum of the top $k$ singular values of the matrix $T=\hat{\Sigma}_{11}^{-1 / 2} \hat{\Sigma}_{12} \hat{\Sigma}_{22}^{-1 / 2}$. If we take $k=d$, then this is exactly the matrix trace norm of $T$; $
\operatorname{corr}(\mathcal{Z}_{1}, \mathcal{Z}_{2})=\|T\|_{\operatorname{tr}}=\operatorname{tr}(T^{\prime} T)^{1 / 2}
$.


Both the networks parameters are updated by computing the gradient  of $\operatorname{corr}(\mathcal{Z}_{1},\mathcal{Z}_{2})$ and update the parameters using  backpropagation. If the singular value decomposition of $T$ is $T=U D V^{\prime}$, then
\begin{equation}\label{tab:canonical_gradient}
\frac{\partial \operatorname{corr}(\mathcal{Z}_{1}, \mathcal{Z}_{2})}{\partial \mathcal{Z}_{1}}=\frac{1}{d-1}(2 \nabla_{11} \bar{\mathcal{Z}}_{1}+\nabla_{12} \bar{\mathcal{Z}}_{2})
\end{equation}

where
$$
\nabla_{12}=\hat{\Sigma}_{11}^{-1 / 2} U V^{\prime} \hat{\Sigma}_{22}^{-1 / 2}
$$
and
$$
\nabla_{11}=-\frac{1}{2} \hat{\Sigma}_{11}^{-1 / 2} U D U^{\prime} \hat{\Sigma}_{11}^{-1 / 2}
$$

\subsection{DCCDI Model}
The meta-training approach of our proposed \textbf{DCCDI} method can be found in Algorithm 1. Our primary focus in this work is to improve the generalization ability of few-shot classification models to unseen domains by learning a prior on the model weights which is suitable for Meta-Fine-tuning during the meta-testing phase on document datasets.  We have also proposed a canonical-correlation-based layer in the model to integrate effectively both the textual and visual modalities of the document images which can be seen as a fine-grained cross-modality alignment task.

\subsubsection{Domain Agnostic Meta-Learning for Document Intelligence}
Our focus in this work is to improve the generalization ability of our meta-trained model to arbitrary unseen document intelligence domains. The Meta-training that incorporates \textbf{CCDI} its described in detail in Algorithm 1.

We aim to train a model that can adapt swiftly to novel unseen classes. This problem setting is often formalized as cross domain few-shot learning. In this proposed approach, the model is meta-trained on a set
of tasks generated using $\mathcal{D}_{source}$, such that the meta-trained model can quickly adapt to new unseen novel tasks using only a small number of examples or trials  generated using $\mathcal{D}_{target}$. In this section, we formally state the problem and present the general form of our algorithm. Similar to Meta-Learning algorithms the proposed algorithm can be subdivided into following phases.\\

\subsubsection{Meta-Training Phase}
We used ResNet-10 as our visual feature extractor or encoder. It have been shown recently that this pre-training process significantly improves the generalization \citep{rusu2018meta,gidaris2018dynamic,lifchitz2019dense}. We pre-train the visual feature encoder on a source dataset (miniImageNet or tiredImageNet) by incorporating a final linear layer.

After the pre-training stage, we start our meta-training process of few-shot classification training stage. First, we train and update the last $k$ layers of the visual feature encoder $\mathnormal{E}$ followed by a linear classifier layer. We minimize the standard cross-entropy loss on the meta-training dataset by using only the support set images as shown in the Stage-1 of Figure \ref{fig:overview}.  After this Stage-1 training process, all the layers of the visual feature encoder block of the model $f_{\theta}$ will be unfrozen.

In the Stage-2 phase, we train the proposed model using the traditional episodic meta-learning paradigm. For each episode a new task $\mathcal{T}_{i}$ is sampled from $\mathcal{D}_{source}$, the model $f_{\theta}$ is trained with $N$ samples present in the support set. The model is then tested on query samples from the same task. The prior parameters of the model $f$ are then updated by considering the test error on the query set. Actually, the test error on sampled tasks $\mathcal{T}_i$ serves as the training error of the meta-learning process. All the parameters in the network are updated using the MAML \citep{finn2017model} first order approach. For this stage-2, we proceed similarly to  \citep{guo2020broader,chen2021self,cai2020cross} which successfully use a metric mapping module to  project the final linear classifier scores onto a metric space which can be used to compare support and query samples, hence increasing the overall accuracy. A graph neural network is used for the Metric-Learning module which is similar to architecture used in few-shot graph neural networks \citep{garcia2018fewshot}.

\subsubsection{Meta-Testing or Meta-Deployment Phase}
At the start of the meta-test phase the first $l$ layers of the visual feature extractor was frozen and the last $k$ layers are left unfrozen. With the main goal of adapting the meta-trained model for the business document domain, we introduce the CCDI module during our deployment phase. During Meta-Testing, a new task $\mathcal{T}_{i}$ is drawn from the $\mathcal{D}_{target}$. The input document images are resized to 224 × 224 then fed into the visual feature blocks. Visual features are extracted for each of the document image present in the support set using the visual feature encoder block from the meta-trained model $\hat{f}_{\theta}$. Similarly for each of the document images, text is extracted using Pytesseract and then fed into pre-trained longformer model \citep{Beltagy2020Longformer} to extract textual embedding features. Both the textual and visual modalities are passed through its corresponding deep Canonical Correlation bock and jointly optimized. Training the canonical co-relation block results in representations that aligns both the modalities (Image and text). The resulting  meta-trained model along with the metric module, which consists of an ensemble of a graph convolution neural network classifier and a linear classifier layer is then trained on this data. Finally both the scores are combined and treated as the final classification scores.

\RestyleAlgo{algoruled}
 \begin{algorithm}[h!]
 {\small
 \caption{DCCDI Meta-Test Protocol}
        \DontPrintSemicolon
        \SetKwInput{kwInput}{Require}
        \SetKwInput{kwInput}{Require}
        \SetKwInput{kwInput}{Require}
        \SetKwInput{kwInput}{Require}
        \SetKwBlock{kwMain}{Meta-Test}{end}
        \SetKwInput{kwOutput}{Output}
        \kwInput{
            $p(\mathcal{T})$  $\leftarrow \mathcal{T}_{1 . . n}$ ; $n$  meta-testing tasks generated from $\mathcal{D}_{target}$ in the form of K shot-N-way classification
            }
        \kwInput{
           $\hat{f}_{\theta} \leftarrow$ meta-trained model
            }  
            
                \kwInput{
          $\hat{f}_{bert} \leftarrow$ pre-trained longformer model
            }  
              \kwInput{
           $g_{\phi_{1}}, g_{\phi_{2}} \leftarrow$ Canonical Corelation Module
            }  
        \kwMain{
        \par
        \vspace{0.25cm}
                 \For{i=1,...,$n$}{
                Sample Task $\mathcal{T}_{i}$ from $p(\mathcal{T})$ \\
                Sample $N$ samples from  support set $\mathcal{S}$ and $M$ samples from query set $\mathcal{Q}$. \\
                Obtain visual, texual modality $\mathcal{Z}_{1}=\textsl{g}_{\phi_{1}} (\hat{f}_{\theta}(\mathcal{S}))$ and $\mathcal{Z}_{2}=\textsl{g}_{\phi_{2}} (\hat{f}_{\theta}(\mathcal{S}))$
                Train the canonical co relation module for 20 steps using canonical loss as shown in Eq \ref{tab:canonical_gradient} $\nabla_{\phi_{1},\phi_{2}} \mathcal{L}_{\mathcal{T}_{i}}(\mathcal{Z}_{1},\mathcal{Z}_{2})$  by following the \\
                Concatenate the visual features and low dimensional textual features found using the trained canonical co-relation module $\hat{g}_{\phi_{1}}, \hat{g}_{\phi_{2}}$\\
                
                Feed the concatenated feature vectors through feature extractor and then
                through the metric learning module\\

                Calculate $\nabla_{\theta} \mathcal{L}_{\mathcal{T}_{i}}(\hat{f}_{ \theta}(\mathcal{Q}))$  using the query set samples\\
                
                Compute adapted parameters with gradient descent and update the model parameters.
                

        }
        \vspace{0.25cm}
        }
        \label{psuedocode}
        }
    \end{algorithm}

\vspace{-0.5cm}
\section{Experiments and Results}

\begin{table*}[ht]
{\small
\centering
\begin{tabular}{cccccccc}
\toprule
     \multirow{3}{*}{\textbf{Baselines}} &   \multirow{3}{*}{\textbf{ Embedding Net }} & \multicolumn{3}{c}{INSR Dataset (5-way)}&\multicolumn{3}{c}{RVL Dataset (5-way)}\\
        & & \textbf{ 5-shot} & \textbf{10-shot} & \textbf{20-shot}  & \textbf{5-shot} & \textbf{10-shot} & \textbf{20-shot}\\
\midrule
ProtoTypical Networks & Conv-4 & 57.97 \% & 62.21  \% & 65.4\% &44.64\% &48.85 \% & 53.42 \% \\
ProtoTypical Networks & Resnet-10 & 50.04 \% & 53.33\% & 52.78\% & 49.50 \% & 52.58 \% & 53.25 \%\\
Relational Networks   & Conv-4 & 55.47  \% &  56.58 \% & 57.27\%  & 41.28  \% & 46.93 \% & 49.10 \%\\
Relational Networks   & ResNet-10 & 32.42  \% & 43.08\% & 46.53\%  & 33.08  \% & 34.68 \% &  37.57\% \\
Matching Networks      & ResNet-10 & 48.53  \% & 50.21\%&58.92\%& 40.57  \% & 44.76 \% & 52.21\%\\
\textbf{DCCDI without textual features}          & ResNet-10& \textbf{65.21} \% & \textbf{70.93}\% & \textbf{77.2}\% & \textbf{60.85} \% & \textbf{66.45} \% & \textbf{70.32} \%\\
\textbf{DCCDI} & ResNet-10 & \textbf{67.82} \% & \textbf{72.79} \% &\textbf{79.78}\% &\textbf{61.76} \% & \textbf{67.40} \% & \textbf{72.93 }\%\\
                \addlinespace[1ex]
            \hline
\bottomrule
\end{tabular}
\caption{\textbf{Few Classification accuracy on the INSR, miniRVL dataset when source domain is miniImagenet.}}
\label{tab:miniImageNet}

\bigskip

\begin{tabular}{cccccccc}
\toprule
     \multirow{3}{*}{\textbf{Baselines}} &   \multirow{3}{*}{\textbf{  Embedding Net }} & \multicolumn{3}{c}{INSR Dataset (5-way)}&\multicolumn{3}{c}{RVL Dataset (5-way)}\\
        & & \textbf{ 5-shot} & \textbf{10-shot} & \textbf{20-shot}  & \textbf{5-shot} & \textbf{10-shot} & \textbf{20-shot}\\
\midrule
ProtoTypical Networks & Conv-4 & 61.09 \% & 61.41  \% & 65.18\%&44.42\% &46.94 \% & 54.50 \% \\
ProtoTypical Networks & Resnet-10 & 52.21 \% & 52.52\% & 53.76\% & 46.86 \% & 46.18 \% & 52.30 \%\\
Relational Networks   & Conv-4 & 50.28  \% & 56.82 \% & 60.29\%& 41.30  \% & 48.61 \% & 47.61 \%\\
Relational Networks  & ResNet-10 &  29.36\%& 39.36  \%  & 54.82\% & 26.58  \% & 29.81 \% & 44.29 \% \\
Matching Networks    & ResNet-10 &40.09 \% & 45.88\%& 48.10   \%&33.68\% & 42.37 \% & 43.92\%\\
\textbf{DCCDI without textual features}      & ResNet-10& \textbf{65.73} \% & \textbf{71.75}\% & \textbf{77.11}\%& \textbf{61.32} \%  & \textbf{66.92 }\% & \textbf{71.04} \%\\
\textbf{DCCDI}  & ResNet-10 & \textbf{66.68} \% & \textbf{74.01}  \% & \textbf{78.85} \% & \textbf{62.05}  \% & \textbf{67.55}  \% & \textbf{72.61}  \%\\
                \addlinespace[1ex]
            \hline
\bottomrule
\end{tabular}
\caption{\textbf{Few Classification accuracy on the INSR, miniRVL dataset when source domain is tieredImageNet.}}
\label{tab:tieredImageNet}
}
\end{table*}

In this section, we first introduce the different document image datasets used in our training and evaluation process, then show quantitative comparisons against other baseline methods in the following sections. The code will be attached as supplementary material and will be released publicly after the conference.

\subsection{Datasets}

Robustness of the proposed approach  has been tested using standardized few-shot classification datasets \textbf{miniImageNet} \citep{ravi2016optimization}, \textbf{tieredImageNet} as the single source domain, and evaluate the trained model on two document domain datasets as target domain. The two training datasets are  MiniImageNet dataset which is a subset of bigger ImageNet derived from ILSVRC- 2012 \citep{russakovsky2015imagenet} which is used in our meta-training process consists of 60,000 images from 100 classes, and each class has 600 images in which 64 classes for training, 16 classes for validation, 20 classes for the test. We have used 64 classes using our meta-training phase for episodic training of the proposed model. The TieredImageNet dataset is another dataset derived from ImageNet, it consists 608 classes in total derived from  34 high-level categories. These categories are split into
20 meta-training, 6 meta-validation  and 8 meta-test , which corresponds to 351 train classes, 97 validation classes and 160 test classes respectively. We have used  351 train classes for the training the model using episodic training.\\

\begin{figure}
\begin{center}
\def\tabularxcolumn#1{m{#1}}
\begin{tabularx}{\linewidth}{@{}cXX@{}}
\begin{tabular}{cc}
\subfloat[Medical Record]{\includegraphics[width=3.5cm,height=3.5cm]{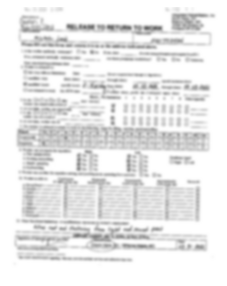}} 
   & \subfloat[Subrogation Letter]{\includegraphics[width=3.5cm,height=3.5cm]{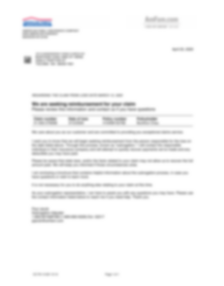}} \\
\end{tabular}
\end{tabularx}
\end{center}
\caption{Examples of documents from the Insurance data set. The document classes (a) Medical Record, (b) Subrogation Letter (blurred for privacy reasons).}
\label{insurancesample}
\end{figure}

We use the following two datasets to evaluate our proposed model. The Insurance company dataset (Fig.\ref{insurancesample}), dubbed as \textbf{INSR} for anonymity contains 5772 document images which spans across 11 categories (few examples are shown in Fig \ref{insurancesample}). Some Categories from the INSR dataset include Medical Bills, Medical Authorizations, Medical Records etc. 

The second dataset is a few-shot learning dataset dubbed as The \textbf{miniRVL} dataset \cite{harley2015evaluation} that has been generated from a larger RVL dataset which consists of 400,000 images which spans across 16 categories. The data relates to  document classification and include Advertisements, Emails among other document types. The \textbf{miniRVL} dataset  consists of16 classes with 1000 images per class and it is designed to keep the inter-class similarity sufficiently high to purposely pose a few-shot learning document classification challenge.


\subsection{Document Pre-processing}
To construct the textual features for each document image, we use PyTesseract to extract all the text present in the document. All the extracted text is then passed through longformer model pre-trained on \textit{longformer-base-4096} \citep{Beltagy2020Longformer} and the textual feature vector is then collected for each image. We have used longformer models due to it's increased capacity to handle documents of thousands of tokens or longer as we want to utilize all the textual information present in the document images.

For all the experiments, we measured performance by computing the average accuracy across 3 independent runs.

\subsection{Implementation Details:} We compared our method to three metric-based learning methods: Matching Networks \citep{vinyals2016matching}, Relation Networks \citep{sung2018learning}, and Prototypical Networks \citep{snell2017prototypical}. We also compared the proposed method using visual and textual features to the proposed method using only visual features. Prior to our meta training phase we also pre-trained our image feature extractor by minimizing the standard supervised cross-entropy loss on the source domain dataset such as miniImageNet or TieredImageNet. This is similar to several recent works \citep{rusu2018meta,gidaris2018dynamic,lifchitz2019dense} that have shown significant improvement in classification accuracy via this method. In all our experiments, we used ResNet-10 model as the backbone of the visual encoder. We also included ablation studies to see the effect of our proposed method for various hyper parameters. 

\subsection{Hyper parameters and infrastructure}
We use easyfsl\footnote{https://github.com/sicara/easy-few-shot-learning},
an open-source deep learning toolkit with a number of implementations of metric-based learning methods. We use a Tesla V100 GPU, consisting of 16GB of memory for all models requiring GPU for train, and use amazon EC2-t2.micro. We use PyTorch 1.9 as the backend framework, and the Pytesseract package for extracting the text from the document images. We used the public architecture implementation from official matching networks  \citep{chen2019closer} and relational networks, and the graph neural network is trained using the official
implementation for few-shot graph convolutional network\footnote{https://github.com/vgsatorras/few-shot-gnn}. We used the canonical correlation block containing two neural networks, their architectures as shown in Table \ref{tab:albation_architectures}. Both the network parameters were optimized together using canonical loss using the RMSprop optimizer with a learning rate of $0.001$\\


\begin{table}
{\small
\begin{tabular}{ccc}
\toprule 
\textbf{Layer} & \textbf{Visual Feature} & \textbf{Text-Feature}\\
\midrule
Input & 512 & 786 \\
1st layer & $1024, \tanh$ & $1024, \mathrm{tanh}$ \\
2nd layer & $1024, \tanh$ & $1024, \tanh$ \\
3rd layer (output) & 20, linear & 20, linear\\
\hline
\bottomrule
\end{tabular}
\caption{Architecture of CCDI Block}
\label{tab:albation_architectures}
}
\end{table}
\subsection{Comparing classification accuracy} 
\subsubsection{MiniImageNet}
Results comparing the baselines to our model on meta-trained on miniImagenet and deployed on document image datasets are shown in Table \ref{tab:miniImageNet}. For 5-shot 5-way, 10-shot 5-way, and 20-shot 5-way, our proposed model outperforms all the existing baselines. As shown in the Table \ref{tab:miniImageNet} all the baseline models, Prototypical Networks, Matching Networks and Relational Networks work well when the embedded model is Conv-4 and the performance degrades rapidly when a Resnet-10 block was used as an embedding model for each metric-based baseline method. As shown, the baseline method which closest performance to our proposed approach Prototypical network which achieves 61.09\% accuracy for (5-shot, 5-way). However, performance doesn't improve much for both the (10-shot, 5-way) and the (20-shot, 5-way) classification. The proposed CCDI Model without the canonical co-relation block achieved an accuracy of 65.73\% and the one with the canonical co-relation block has achieved an state-of-the-art accuracy of 66.68\%  which shows the significance of proposed method during the meta-testing phase.

Our experiments followed the same setup described above for meta-testing on the open source RVL dataset. Results are shown in Table \ref{tab:tieredImageNet}. However the miniRVL dataset classification task is more challenging as it contains many documents that doesn't follow specific layout structures within classes. 
As shown Table \ref{tab:tieredImageNet} (b),  the proposed method also outperforms the rest of the baseline methods and achieves state-of-the-art performance on this dataset by a large margin of up to 12\% when compared to its closer competitor: Relational Networks. 

\subsubsection{TieredImageNet} To test the effectiveness of the proposed approach when meta-trained on a bigger domain, we repeated the experiments of the previous section.
Results are shown in Table \ref{tab:tieredImageNet}. On the INSR dataset, the models PrototTypical Networks(Conv-4), PrototTypical Networks(ResNet10), Relational Networks(Conv-4),Relational Networks(ResNet10)and Matching Networks(ResNet10)
achieved mean accuracies of [61.09\%, 52.21\%, 50.28 \%, 39.36\%, 48.10\%], respectively for (5-way, 5-shot) classification. In comparison, our proposed method achieved an accuracy of 
66.68\%. PrototTypical Networks (Con-4) came in second, with an accuracy of
only 61.09\% , followed by Relational Networks Networks(Con-4) with an accuracy of
50.28\%. Finally, we found that our proposed model outperform those existing baselines
in all the scenarios of  (5-way, 5-shot), (10-way, 5-shot) , (20-way, 5-shot) and results can be found in the Table \ref{tab:tieredImageNet}.

\begin{table}[ht]
\centering
{\small
\begin{tabular}{cc} 
\toprule
\textbf{Vector Size} & \textbf{Accuracy}\\
\midrule
    15 &  64.75\% \\
    20 & \textbf{65.92\% } \\
    25& 64.88\%  \\
\hline
\bottomrule
\end{tabular}
\caption{Effect on Output Vector dimension on Accuracy}
\label{tab:albation_vectorsize}
}
\end{table}
\vspace{-0.6cm}
\begin{table}
\centering
{\small
\begin{tabular}{cc}
\toprule 
No of Aug Images & Accuracy \\
\midrule
0- Augmented Images & \textbf{66.68\% } \\
5- Augmented Images  & 65.44\% \\
10- Augmented Images & 65.39\%   \\
15- Augmented Images  & 64.59\%  \\
\hline
\bottomrule
\end{tabular}
\caption{Evaluation Augmentation Results on INSR Dataset when meta-trained on TieredImagenet}
\label{tab:ablation_augmentaion}
}
\end{table}
\subsection{Ablation Studies}\label{tab:ablationstudies}
 Several ablation studies has been done to understand the significance and impact of the different hyperparameters on the proposed model. To evaluate the effectiveness of our proposed canonical correlation module for combining the textual and image features for document image classification, we first investigate the performance of the canonical correlation  module for different output dimensions for $\mathcal{Z}$ based on the textual content and visual features.
We did grid search the output dimensions from the set of [10,15,20,25] and results are shown in Table \ref{tab:albation_vectorsize}. The vector of size 20 gave the best performance results for different experiments. 

We also evaluate the effectiveness our proposed approach for aligning the textual and image features instead of directly concatenating the textual and image features for document classification. As it is shown in Table \ref{tab:albation_concatvs_ccdi}, the DCCDI method achieves higher accuracies than the widely used concatenation of the image and textual features. 
We also performed an ablation study to see the effect of data augmentation on the generalization of performance on unseen data during the meta-test phase. As it is shown in various studies, data augmentation helps in the process of generalization for visual tasks which provide view-point invariance for each visual image. For this, we sampled additional images from the
support set, and perform jitter, random crops, and horizontal flips on a randomized basis. As we can see in Table \ref{tab:ablation_augmentaion} applying data augmentation does not have a positive effect on the solution. One reason for this might be that document images contain specific structure in them.

\begin{table}[ht]
{\small
\centering
\begin{tabular}{ccc} 
\toprule
\textbf{ Method} &\textbf{5-Shot} &\textbf{20-Shot}\\
\midrule
Text-Concat & 66.64\%  &  78.51\%  \\
DCCDI & \textbf{67.8\% } & \textbf{79.78\% } \\
\hline
\bottomrule
\end{tabular}
\caption{Effect on CCDI block vs text-concatenation}
\label{tab:albation_concatvs_ccdi}
}
\end{table}

\section{Conclusion and Future Work}
In many companies, millions of unlabeled documents containing information relevant to many business-related workflows have to be processed to be classified or / and to extract key information. Unfortunately, a large percentage of these documents consists of unstructured formats in the form of images and PDF documents. Examples of these types of documents include: medical bills, attorney letters, contracts, bank statements and personal checks. This process automation it usually refer as "Document intelligence" and relies on the use of models that combine both the image and text information to classify, categorize, and extract the relevant information. However, the labeled data needed by traditional learning approaches maybe too expensive and taxing on business experts and hence not practical in real-world industry settings. Hence, a few-shot learning pipeline is highly desired.

In this work, we proposed a novel method for few-shot document image classification under domain shift for semi-structured business documents, using the canonical correlation block to align extracted text and image feature vectors. We evaluate our work by extensive comparisons with existing  methods on two datasets.
We rigorously benchmarked our method against the state-of-the-art few-shot computer vision models on both an insurance process derived dataset and the miniRVL dataset. We also presented the different ablation studies to show the effectiveness of the proposed method. The results showed our method consistently performed better than existing baselines on few-shot classification tasks. For future work, we would like to further explore more effective document representations including more sophisticated graph representations , or jointly trained layouts \cite{mandivarapu2021efficient} and future directions of implementing the continual learning in document classification \cite{10.3389/frai.2020.00019,blake1,jay2,DJay}.

\bibliography{aaai22.bib}


\end{document}